\documentclass[10pt,twocolumn,letterpaper]{article}

\usepackage[pagenumbers]{wacv} 

\definecolor{wacvblue}{rgb}{0.21,0.49,0.74}
\usepackage[pagebackref,breaklinks,colorlinks,allcolors=wacvblue]{hyperref}
\usepackage{graphicx}
\usepackage{amsmath}
\usepackage{amssymb}
\usepackage{booktabs}
\usepackage{adjustbox}
\usepackage{mathtools}

\newcommand{\nbf}[1]{\noindent\textbf{#1}}

\def\wacvPaperID{121} 
\def\confName{WACV}
\def\confYear{2026}

\title{High-Rate Mixout: Revisiting Mixout for Robust Domain Generalization}

\author{Masih Aminbeidokhti\thanks{Correspondence to: masih.aminbeidokhti.1@ens.etsmtl.ca. Code at \href{https://github.com/Masseeh/HR-Mixout}{https://github.com/Masseeh/HR-Mixout}.} \quad
Heitor Rapela Medeiros \quad
Srikanth Muralidharan \and
Eric Granger \quad
Marco Pedersoli\\
\emph{École de technologie supérieure, Montreal, Canada}
}

\begin{document}
\maketitle

\begin{abstract}
Ensembling fine-tuned models initialized from powerful pre-trained weights is a common strategy to improve robustness under distribution shifts, but it comes with substantial computational costs due to the need to train and store multiple models. Dropout offers a lightweight alternative by simulating ensembles through random neuron deactivation; however, when applied to pre-trained models, it tends to over-regularize and disrupt critical representations necessary for generalization. In this work, we investigate Mixout, a stochastic regularization technique that provides an alternative to Dropout for domain generalization. Rather than deactivating neurons, Mixout mitigates overfitting by probabilistically swapping a subset of fine-tuned weights with their pre-trained counterparts during training, thereby maintaining a balance between adaptation and retention of prior knowledge. Our study reveals that achieving strong performance with Mixout on domain generalization benchmarks requires a notably high masking probability of 0.9 for ViTs and 0.8 for ResNets. While this may seem like a simple adjustment, it yields two key advantages for domain generalization: (1) higher masking rates more strongly penalize deviations from the pre-trained parameters, promoting better generalization to unseen domains; and (2) high-rate masking substantially reduces computational overhead, cutting gradient computation by up to 45\% and gradient memory usage by up to 90\%. Experiments across five domain generalization benchmarks—PACS, VLCS, OfficeHome, TerraIncognita, and DomainNet—using ResNet and ViT architectures show that our approach, High-rate Mixout, achieves out-of-domain accuracy comparable to ensemble-based methods while significantly reducing training costs.
\end{abstract}

\section{Introduction}

Deep neural networks, such as Vision Transformers (ViTs)~\cite{dosovitskiy2020image} and Convolutional Neural Networks (CNNs)~\cite{lecun1989backpropagation} have driven numerous breakthroughs across computer vision~\cite{he2016deep} and beyond~\cite{dong2021survey}. However, their performance can degrade significantly when there is a substantial distribution shift between the training and test data~\cite{koh2021wilds}. To address this challenge, domain generalization (DG) focuses on developing methods that train models on a limited number of source domains while enabling them to generalize effectively to distinct, unseen target domains during deployment~\cite{zhou2022domain}.

Among state-of-the-art DG techniques, leveraging pre-trained models through ensembling~\cite{lakshminarayanan2017simple} and weight averaging~\cite{wortsman2022model} stand out in terms of their performance on diverse DG benchmarks under fair evaluation protocols in realistic and synthetic settings~\cite{koh2021wilds,gulrajani2020search,rame2022diverse,arpit2022ensemble}. Ensembling combines predictions from multiple fine-tuned models, whereas weight averaging merges the parameters of several models into a single one before making predictions. These approaches can significantly outperform other DG methods by leveraging the diversity among multiple models~\cite{rame2022diverse,wortsman2022model}. However, achieving such outstanding performance necessitates numerous training sessions with varied hyperparameters and initializations~\cite{wenzel2020hyperparameter}. This requirement can become prohibitively expensive, especially with large-scale models and datasets.

\begin{figure*}[htp]
    \centering
    \includegraphics[width=1.0\linewidth]{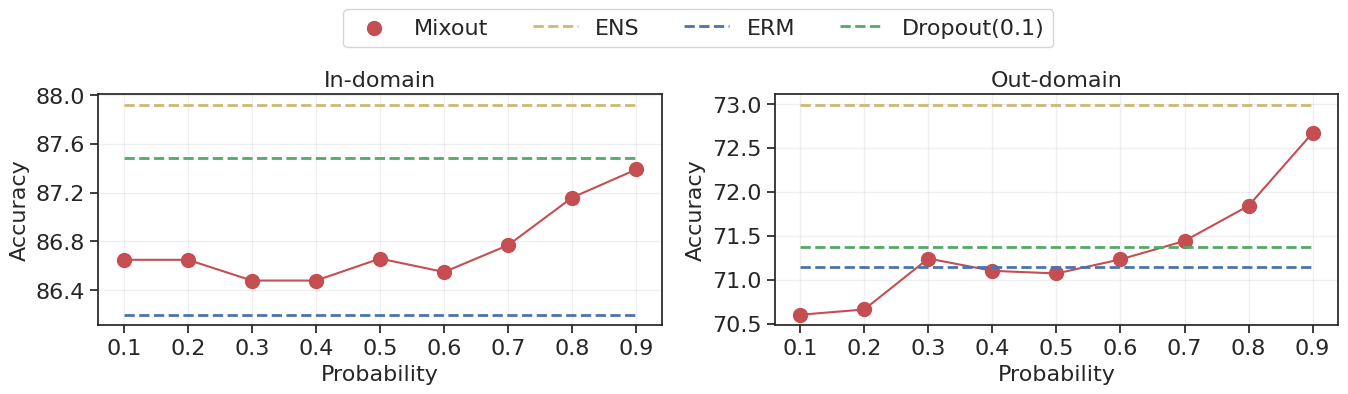}
    \caption{\textbf{Comparison between in-domain (left) and out-of-domain (right) accuracy of Mixout and Dropout on OfficeHome dataset with ViT-S/16 architecture.} While for in-domain the two approaches reach a similar best accuracy, for out-domain the ability of Mixout to preserve the knowledge of the pre-trained model leads to accuracies comparable to ensemble models with a single training. We show Dropout results for a probability of 0.1, as performance rapidly declines to zero for higher probabilities in both cases. Mixout at the probability of 0.0 is equivalent to ERM.}
    \label{fig:vit_iid_ood}
\end{figure*}

A promising approach to approximating ensembles with a single model is Dropout~\cite{srivastava2014dropout}, which trains diverse sub-networks~\cite{bachman2014learning} by randomly deactivating neurons. However, when applied to pre-trained models, Dropout imposes excessive regularization, rapidly erasing crucial pre-trained knowledge essential for generalization to new domains. In this work, we investigate the use of Mixout~\cite{lee2019mixout} as a replacement for Dropout in domain generalization. Mixout dynamically swaps portions of the parameters between the frozen pre-trained model and the learnable weights on the entire network. As shown in Fig.~\ref{fig:vit_iid_ood} (left) and as already described in~\citet{lee2019mixout}, when the training and testing data are drawn from the same distribution, Mixout and Dropout perform similarly, but Mixout can work at much higher rates. 

Our key finding is that under domain shifts (Fig.~\ref{fig:vit_iid_ood}, right), applying Mixout at high rates substantially improves out-of-domain performance, achieving accuracy comparable to ensemble-based methods but at significantly lower computational cost during training. This high-rate application brings three main advantages:
First, unlike static masking approaches~\cite{sung2021training}, Mixout dynamically explores diverse subnetworks during training, encouraging richer and more robust representations.
Second, increasing the masking probability more strongly penalizes deviations from the pre-trained weights, which improves generalization to unseen domains~\cite{wortsman2022robust}.
Third, by updating only a subset of parameters per step, high-rate Mixout reduces computational overhead, cutting gradient computation by up to 45\% and gradient memory usage by up to 90

The contributions of this paper are summarized as follows:
\nbf{(1)} We identify and establish that high-rate Mixout—using aggressive parameter swapping (0.9 for ViTs, 0.8 for ResNets)—is critical for achieving strong domain generalization performance.
\nbf{(2)} We extend Mixout to convolutional architectures by swapping entire convolutional filters rather than individual weights, addressing spatial correlations that limit regularization effectiveness.
\nbf{(3)} We provide extensive empirical validation on five standard domain generalization benchmarks—PACS, VLCS, OfficeHome, TerraIncognita, and DomainNet—showing that High-rate Mixout achieves out-of-domain accuracy comparable to ensemble-based approaches while greatly reducing training costs.

\section{Related Works}

\subsection{Domain Generalization} 
DANN~\cite{ganin2016domain} uses adversarial networks to align feature distributions across domains, inspiring regularization methods like maximum mean discrepancy minimization~\cite{li2018domain}, conditional distribution invariance~\cite{li2018deep,albuquerque2019generalizing}, and covariance alignment~\cite{sun2016deep}. Instead of feature-space regularization, IRM~\cite{arjovsky2019invariant} enforces a shared optimal classifier across domains, while Fish~\cite{shi2021gradient} and IGA~\cite{koyama2020out} use gradient alignment for consistency. GroupDRO~\cite{sagawa2019distributionally} minimizes worst-case loss by prioritizing challenging samples. Meta-learning~\cite{bui2021exploiting} adapts parameters to new domains.
Data augmentation is another powerful strategy to enhance DG by expanding the diversity of the training dataset. Techniques such as RandAugment~\cite{cubuk2020randaugment}, TrivialAugment~\cite{muller2021trivialaugment}, AugMix~\cite{hendrycks2019augmix}, and MixUp~\cite{zhang2017mixup} create robust models by introducing variability in the training data. Beyond heuristic augmentations, some methods leverage domain meta-data to learn challenging and diverse transformations~\cite{yan2020improve,aminbeidokhti2024domain} or synthesize novel domains through style mixing~\cite{zhou2021domain} or generative models~\cite{goel2020model}.
Although the conventional DG algorithms are developed with strong theoretical foundations, these methods
have not consistently outperformed Empirical Risk Minimization (ERM) on fair model selection criteria~\cite{gulrajani2020search,koh2021wilds}. In contrast, Ensemble-based approaches show a substantial performance improvement over the ERM and other conventional DG algorithms by a large margin~\cite{wortsman2022model,rame2023model}.

\subsection{Ensemble And Weight Averaging}
Ensembling deep networks by combining logits from multiple models has been shown to consistently improve robustness~\cite{lakshminarayanan2017simple,fort2019deep} and has been successfully applied to domain generalization~\cite{rame2022diverse,arpit2022ensemble}. This effectiveness stems from the diversity among ensemble members, which arises primarily from the randomness of the learning process~\cite{dietterich2000ensemble,wenzel2020hyperparameter}. However, ensembling requires multiple forward passes per input during inference, making it impractical for real-world deployment. Model soups~\cite{wortsman2022model} address this limitation by combining ensemble models in the weight space rather than at the prediction stage, effectively collapsing multiple models into a single one, streamlining inference. Despite this advantage, deep ensembles and model soups still require training multiple networks, resulting in significant computational overhead. In this study, we show that applying Mixout with a high masking probability during fine-tuning achieves ensemble-level performance without the need to train or store multiple models. Finally, implicit ensemble methods approximate deep ensembles by combining model weights collected throughout a single training run~\cite{cha2021swad,izmailov2018averaging,arpit2022ensemble}. As we show later, these methods are fully compatible with Mixout and can sometimes further improve final performance. 

\subsection{Dropout}
Dropout~\cite{srivastava2014dropout} is a widely used regularization technique that improves in-domain performance by randomly deactivating a subset of neuron activations during training. This stochastic masking injects noise into the network, effectively training an implicit ensemble of subnetworks and enhancing generalization. Building on this idea, various methods have introduced randomness at different structural levels, including entire layers~\cite{huang2016deep,larsson2016fractalnet}, channels~\cite{tompson2015efficient}, neuron connections~\cite{wan2013regularization}, or contiguous regions of feature maps~\cite{ghiasi2018dropblock}, all aiming to boost in-domain generalization. While these techniques are typically applied when training models from scratch, applying Dropout directly to pre-trained models can disrupt critical learned representations and undermine the benefits of pre-training~\cite{zhang2024fine}. To address this, Mixout~\cite{lee2019mixout} offers a more stable alternative by stochastically blending pre-trained and fine-tuned parameters. In this work, we show that for Mixout to be effective in domain generalization, the probability of weight swapping must be significantly high. Furthermore, while unstructured Mixout masking works well for ViT backbones with MLP layers, its effectiveness diminishes in convolutional networks due to strong spatial correlations across features~\cite{ghiasi2018dropblock}. To overcome this limitation, we propose a structured approach: swapping entire convolutional kernels within ResNet layers, which preserves spatial coherence and enhances regularization effectiveness.

\subsection{Parameter Efficient Fine-tuning}
Parameter-Efficient Fine-Tuning (PEFT) methods aim to reduce the substantial cost of fine-tuning large-scale models by updating only a fix small subset of parameters relative to the full model~\cite{han2024parameter}. Among the most widely used PEFT approaches, Low-Rank Adaptation (LoRA)~\cite{hu2021lora} stands out for introducing no additional inference overhead; it approximates weight updates during fine-tuning using low-rank matrices that are merged with the pre-trained weights prior to inference. In contrast to LoRA, our approach updates all model parameters but reduces training cost by stochastically updating only a small fraction of them at each iteration, as we use Mixout with high masking probability. This selective parameter update leads to substantial reductions in gradient computation and memory usage during backpropagation, providing computational efficiency while maintaining full-model flexibility.

\section{Domain Generalization with Mixout}

DG aims to train models that generalize to unseen target domains using only data from source domains. Standard neural networks trained with ERM~\cite{vapnik1991principles} often struggle to generalize when the target domain is far from the source domains~\cite{gulrajani2020search}. Model ensembling, which combines multiple networks either in the output space or weight space, has proven effective in improving out-of-distribution generalization~\cite{wortsman2022model,rame2022diverse}. However, the computational demands of training multiple networks make it impractical to apply to large-scale models. We aim to replicate the robustness of model ensembling within a single training run. In this section, we first review the DG problem formulation under ERM. We then present the High-rate Mixout algorithm in detail. Finally, we examine the relationship between Mixout and model ensembling.

\subsection{Problem Formulation}

We study the problem of multi-source domain generalization for classification. During training, we assume access to $I$ datasets, each containing examples for the same task but collected under different domains or environments. Let $\mathcal{D} = \{ D_i \}_{i=1}^{I}$ denote the set of training domains, where each $D_i$ is a distribution over the input space $\mathcal{X}$. From each domain, we observe $N_i$ training data points, each comprising an input $x$ and a corresponding target label $y$, i.e., $\{(x_j^i, y_j^i)\}_{j=1}^{N_i} \sim D_i$. We denote the size of the full training data by $N$. Similarly, we define a set of target domains $\mathcal{T} = \{ T_i \}_{i=1}^{T},$ which are disjoint from the training domains $\mathcal{T} \cap \mathcal{D} = \emptyset$. The goal of DG is to learn model parameters $\theta \in \Theta$ using training domains $\mathcal{D}$ such that the model generalizes well to the unseen target domain(s) $\mathcal{T}$. ERM~\cite{vapnik1991principles} defines the population risk over the target domains as
\begin{equation} 
\mathcal{R}_\mathcal{T}(\theta) = \mathbb{E}_{(x,y) \sim T_i} \Bigl[ \ell\bigl(f(x; \theta), y\bigr) \Bigr],
\end{equation}
where $f(\cdot;\theta)$ is a model parameterized by $\theta$, and $\ell(\cdot,\cdot)$ denotes the cross-entropy loss function. However, since target domains are unavailable during training, we instead minimize the empirical risk over the observed training domains
\begin{equation} 
\hat{\mathcal{R}}^{\text{ERM}}_\mathcal{D}(\theta) = \frac{1}{IN} \sum_{i=1}^{I} \sum_{n=1}^{N_i} \ell\bigl(f(x_n^i; \theta), y_n^i\bigr).
\end{equation}
Due to domain shifts, optimizing $\theta^* = \arg\min_{\theta} \hat{\mathcal{R}}^{\text{ERM}}_\mathcal{D}(\theta)$ with standard gradient descent often yields multiple solutions that achieve similar training losses yet exhibit significantly different generalization performances on $\mathcal{R}_\mathcal{T}(\theta)$. Ensemble-based methods~\cite{lakshminarayanan2017simple,rame2022diverse,arpit2022ensemble,wortsman2022model} address this issue by exploring and combining multiple basins of the loss landscape~\cite{fort2019deep}; however, they require training numerous independent models. Our goal is to propose a method that retains the robustness benefits of ensembling while requiring only a single training run.

\subsection{High-rate Mixout}

Consider $\theta$ and $\theta_0$ as the fine-tuned and pre-trained parameters, respectively. During training with Mixout, at each iteration, a binary mask is generated, where each element is independently sampled from $\text{Ber}(p)$, a Bernoulli distribution with masking probability of $p$. According to the mask, the corresponding model parameters are then swapped with their pre-trained counterparts, effectively freezing these parameters. This process yields the following empirical risk:
\begin{align}
    \hat{\mathcal{R}}_\mathcal{D}^{\text{Mixout}}(\theta) &= \frac{1}{IN} \sum_{i=1}^{I} \sum_{n=1}^{N_i} \Big( \mathbb{E}_{\xi \sim \text{Ber}(p)} \Bigl[ \ell\bigl(f(x^i_n; \theta_\xi), y^i_n\bigr) \Bigr] \Bigr), \\
    \text{s.t.} \quad \theta_\xi &=  \theta_0 \odot (\mathbf{1} - \xi) + \theta \odot \xi, \label{eq:theta_bar}
\end{align}
\noindent where $\odot$ denotes the element-wise product, $\mathbf{1}$ is the identity vector, $\xi \sim \text{Ber}(p)$ is a binary mask realization and $\theta_\xi$ denotes the new parameters after substituting the pre-trained values. 

During training, we optimize the stochastic version of Mixout, approximating the inner expectation using a single mask per mini-batch. At inference time, to fully leverage the networks parameters, we switch to a deterministic formulation by computing the expectation over all possible masks. Since explicitly averaging across many mask realizations would require numerous forward passes, we instead adopt a weight scaling approximation, which preserves performance while requiring only a single forward pass, matching the computational efficiency of ERM baselines~\cite{srivastava2014dropout}. Specifically, we scale and shift the parameters from Eq.\eqref{eq:theta_bar} by $\frac{1}{p}$ and $- \theta_0 \cdot (1 - p)$, respectively~\cite{lee2019mixout}.

As shown in our experiments, achieving high performance on domain generalization benchmarks requires a high masking probability of 0.9 for ViTs and 0.8 for ResNets. Applying Mixout at this rate offers three key advantages in out-of-domain scenarios. First, unlike static masking approaches~\cite{sung2021training}, Mixout continuously uncovers new subnetworks throughout training, leading to more diverse representations. Second, as the masking probability goes higher, the deviation from the pre-trained parameters is penalized more, which in effect helps to foster better generalization to unseen domains~\cite{wortsman2022robust}. Lastly, masking at a high rate significantly reduces computational overhead, cutting gradient computation by up to 45\% and gradient memory usage by up to 90\%. 
\begin{figure}[t]
    \centering
    \begin{subfigure}[t]{0.23\textwidth}
     \centering
     \includegraphics[width=0.8\textwidth]{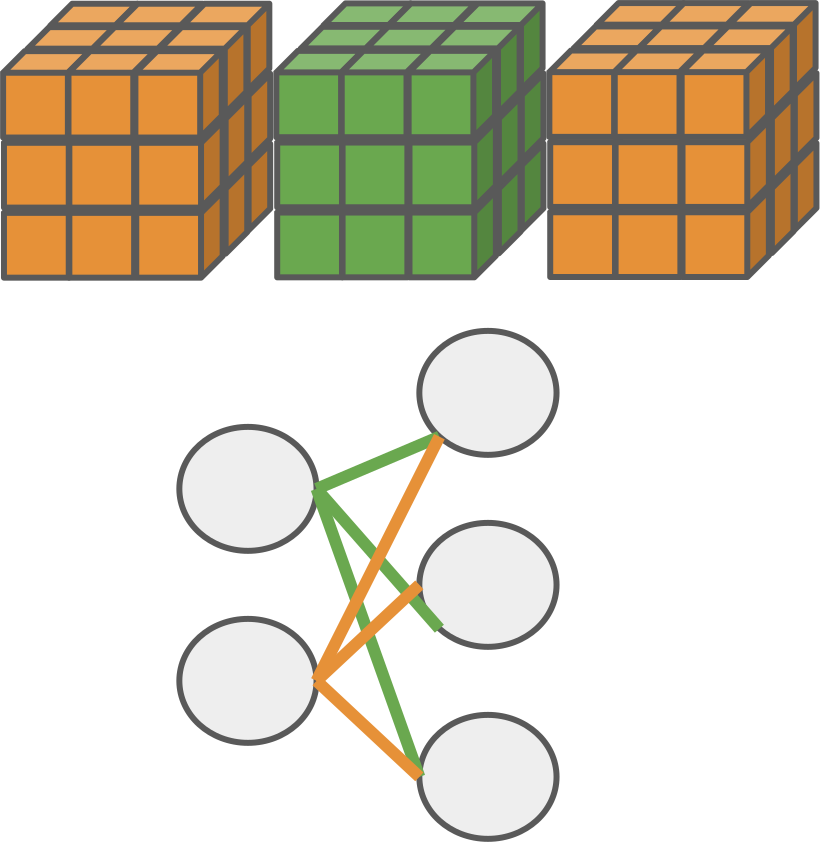}
     \caption{Structural masking.}
     \label{fig:main_a}
    \end{subfigure}
    \hfill
    \begin{subfigure}[t]{0.23\textwidth}
        \centering
        \includegraphics[width=0.8\textwidth]{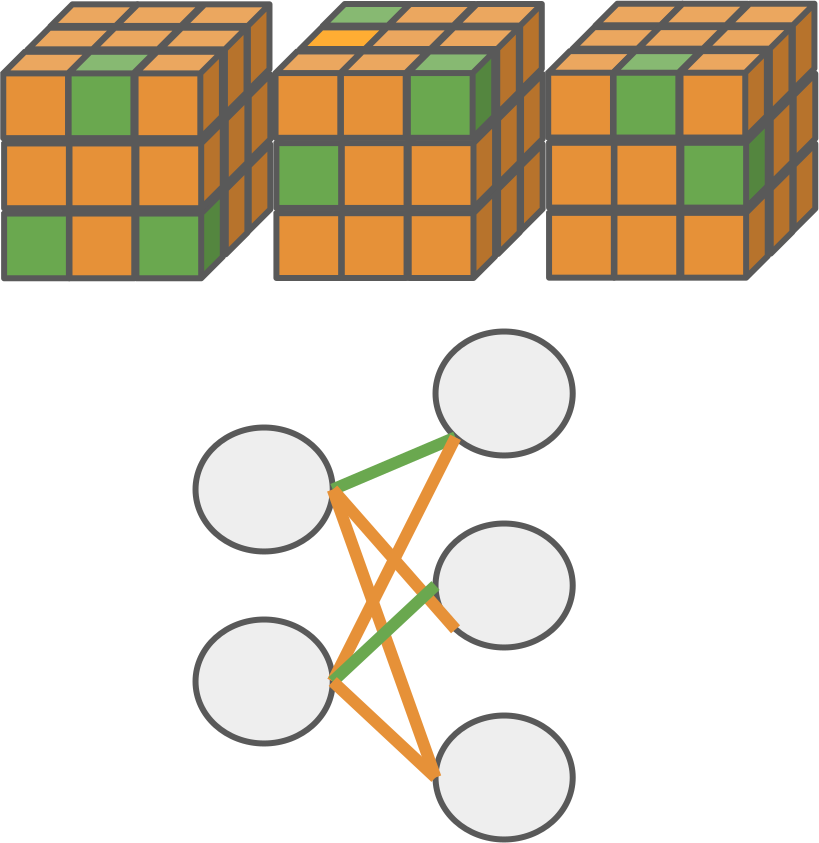}
        \caption{Unstructured masking.}
        \label{fig:main_b}
    \end{subfigure}
    \caption{\textbf{Difference between structural (a) and unstructured (b) Mixout.} The top shows convolutional filters, while the bottom shows neurons.}
    \label{fig:str_mixout}
\end{figure}

\nbf{MLP and Convolutional layers.} For ViTs, we apply Mixout to all MLP layers within both the attention and fully connected blocks. While randomly swapping parameters without structure can be effective for MLP~\cite{lee2019mixout}, its efficacy diminishes when applied to convolutional layers due to the strong spatial correlations among features~\cite{ghiasi2018dropblock}. These correlations allow significant information about the input to propagate through the network even when some parameters are swapped, which can lead to overfitting~\cite{ghiasi2018dropblock}. This highlights the need for a more structured and spatially aware approach to swapping to regularize convolutional networks effectively. To address this, we introduce a variant of Mixout that randomly swaps entire convolutional kernels. We apply this variant to all convolutional layers in ResNet50. Figure \ref{fig:str_mixout} illustrates the difference between structured and unstructured Mixout for both CNN and MLP layers. 

\subsection{Connection to Ensembling}
In practice, Deep ensembles~\cite{lakshminarayanan2017simple} and Model soups~\cite{wortsman2022model} train $K$ networks from different initializations; at inference, deep ensembles average their outputs (requiring $K$ forward passes), whereas model soups collapse the ensemble into one network by averaging parameters, so inference costs only a single pass. Here, we demonstrate the equivalence of Mixout with an ensemble of subnetworks induced by the Mixout process. 

Let $f(\,\cdot\,;\theta)$ be a twice-continuously differentiable function with respect to $\theta$. With Mixout, each mini-batch uses an independent binary mask, yielding a family of masked parameter vectors $\{\theta_{\xi_k}\}_{k=1}^{K}$ with expected value of:
\begin{equation}    
\bar{\theta}
\;=\;
\mathbb{E}[\theta_{\xi_k}]
=
\theta_{0} + p\bigl(\theta-\theta_{0}\bigr).
\end{equation}
We denote $\Delta_k = \theta_{\xi_k} - \bar{\theta}$. Note that $\frac{1}{K} \sum_{k=1}^K \Delta_k = 0$. Averaging the subnetworks predictions gives:
\begin{equation*}
\bar{f} = \frac{1}{K}\sum_{k=1}^{K} f\!\bigl(x;\theta_{\xi_k}\bigr).
\end{equation*}
A first-order Taylor expansion of $f$ around $\bar{\theta}$ yields
\begin{equation}\label{eq:taylor_appx}
f\bigl(x;\theta_{\xi_k}\bigr)
=
f\bigl(x;\bar{\theta}\bigr)
+
\nabla f\bigl(x;\bar{\theta}\bigr)^{\top}\Delta_{k}
+
O\bigl(\lVert\Delta_{k}\rVert^{2}\bigr).
\end{equation}
Taking the expectation from both sides of the Eq.~\ref{eq:taylor_appx} over mask realizations $\xi_k$ cancels the linear term, hence:
\begin{equation}
\bar{f}-f\bigl(x;\bar{\theta}\bigr)
=
O\Bigl(\max_{k}\lVert\Delta_{k}\rVert^{2}\Bigr).
\end{equation}
During fine-tuning, Mixout regularizes the weights to be close to initialization, therefore
$\lVert\Delta_{k}\rVert \approx \sqrt{p(1-p)}\lVert\theta-\theta_{0}\rVert$ remains small~\cite{lee2019mixout}.  
Hence, the deterministic Mixout weights $\bar{\theta}$ serve as a surrogate for the full ensemble over masks, delivering ensemble-level robustness with only one forward pass and one training run.

\section{Experiments} \label{experiments}
Following the DomainBed benchmark~\cite{gulrajani2020search, cha2021swad}, we evaluate our method on five diverse datasets. PACS~\cite{li2017deeper}, VLCS~\cite{fang2013unbiased}, Office Home~\cite{venkateswara2017deep}, TerraIncognita~\cite{beery2018recognition} and DomainNet~\cite{peng2019moment}. We report out-of-domain accuracies for each domain and their average using a leave-one-out cross-validation method (please refer to the appendix for more details on datasets and training configurations). 

We evaluate our method on ResNet50 and ViT-16/S architectures, comparing our approach against ERM, deep ensembles (ENS)~\cite{lakshminarayanan2017simple}, and weight averaging (DiWA)~\cite{cha2021swad,rame2022diverse}—all of which outperform ERM and generally SOTA on DG benchmarks but require training multiple models. To further extend our evaluation on ResNet50, we also include the original Mixout recipe~\cite{lee2019mixout}, which does not use any structure during weight swaps, and CORAL~\cite{sun2016deep}, the strongest domain invariance learning method. Additionally, we re-run the approach from~\citet{zhang2024fine} under the same conditions and configurations as the other models and report its results under ``Large Dropout". For ensemble-based approaches, we combine 18 models from the hyperparameter search phase~\cite{rame2022diverse}, either in weight space (DiWA) or output prediction space (ENS). We report the average performance and standard error for each method over three random seeds. We also report the computation overhead required by each method during inference and training relative to the ERM baseline.

\begin{table*}[pt]
    \centering
   \caption{\textbf{Out-of-domain accuracy on five DG benchmarks from DomainBed.} Cost(T) and Cost(I) denote the computation overhead required by each method during training and inference, respectively, relative to the ERM baseline. Average accuracy and standard error are reported from three trials.}
    \begin{adjustbox}{width=1.0\textwidth}
    \begin{tabular}{lcc|ccccccc}
        \toprule
        {Method} & {Cost(T)} & {Cost(I)} & {PACS} & {VLCS} & {OH} & {TI} & {DN} & {\textbf{Avg.}} \\

        \midrule
        \multicolumn{9}{c}{\textbf{ResNet50}} \\
        \midrule
        \multicolumn{9}{l}{\textbf{Multi-run training (18 Models)}} \\
        \midrule

        ENS~\cite{lakshminarayanan2017simple} &
        18$\times$ &
        18$\times$ &
        89.05  &
        80.03  &
        71.77  &
        54.10  &
        49.11  &
        68.80    \\

        DiWA~\cite{rame2022diverse} &
        18$\times$ &
        1$\times$ &
        89.21  &
        79.83  &
        71.74  &
        55.68  &
        48.40  &
        \underline{68.96}    \\

        \midrule
        \multicolumn{9}{l}{\textbf{Single-run training}} \\
        \midrule

        CORAL~\cite{sun2016deep} &
        1$\times$ &
        1$\times$ &
        87.40$\scriptstyle\pm 0.67$  &
        80.01$\scriptstyle\pm 0.52$  &
        71.23$\scriptstyle\pm 0.34$  &
        50.61$\scriptstyle\pm 1.48$  &
        47.33$\scriptstyle\pm 0.29$  &
        67.32$\scriptstyle\pm 0.66$    \\

        Large Dropout~\cite{zhang2024fine} &
        1$\times$ &
        1$\times$ &
        87.14$\scriptstyle\pm 0.62$  &
        79.31$\scriptstyle\pm 0.43$  &
        70.66$\scriptstyle\pm 0.38$  &
        52.27$\scriptstyle\pm 1.55$  &
        47.36$\scriptstyle\pm 0.17$  &
        67.35$\scriptstyle\pm 0.63$    \\  

        ERM~\cite{vapnik1991principles} &
        1$\times$ &
        1$\times$ &
        87.66$\scriptstyle\pm 0.71$  &
        79.64$\scriptstyle\pm 0.43$  &
        70.46$\scriptstyle\pm 0.70$  &
        52.62$\scriptstyle\pm 2.33$  &
        48.48$\scriptstyle\pm 0.48$  &
        67.77$\scriptstyle\pm 0.93$    \\

        \textbf{High-rate Mixout (Ours)} &
        .73$\times$ &
        1$\times$ &
        87.94$\scriptstyle\pm 0.87$  &
        79.40$\scriptstyle\pm 0.39$  &
        72.14$\scriptstyle\pm 0.30$  &
        58.42$\scriptstyle\pm 0.66$  &
        47.69$\scriptstyle\pm 0.26$  &
        \textbf{69.12$\scriptstyle\pm 0.50$}    \\

        \midrule
        \multicolumn{9}{c}{\textbf{ViT-S/16}} \\
        \midrule
        \multicolumn{9}{l}{\textbf{Multi-run training (18 Models)}} \\
        \midrule

        ENS~\cite{lakshminarayanan2017simple} &
        18$\times$ &
        18$\times$ &
        87.89  &
        80.96  &
        72.99  &
        43.86  &
        49.24  &
        \underline{67.00}    \\

        DiWA~\cite{rame2022diverse} &
        18$\times$ &
        1$\times$ &
        88.38  &
        80.52  &
        72.95  &
        45.09  &
        48.38  &
        \textbf{67.06}    \\

        \midrule
        \multicolumn{9}{l}{\textbf{Single-run training}} \\
        \midrule

        ERM~\cite{vapnik1991principles} &
        1$\times$ &
        1$\times$ &
        86.04$\scriptstyle\pm 0.68$  &
        79.83$\scriptstyle\pm 0.36$  &
        71.14$\scriptstyle\pm 0.38$  &
        42.30$\scriptstyle\pm 1.20$  &
        47.49$\scriptstyle\pm 0.20$  &
        65.36$\scriptstyle\pm 0.56$    \\

        \textbf{High-rate Mixout (Ours)} &
        .69$\times$ &
        1$\times$ &
        86.79$\scriptstyle\pm 0.52$  &
        79.27$\scriptstyle\pm 0.42$  &
        72.58$\scriptstyle\pm 0.18$  &
        45.51$\scriptstyle\pm 0.56$  &
        47.45$\scriptstyle\pm 0.22$  &
        66.32$\scriptstyle\pm 0.23$    \\

        \bottomrule
    \end{tabular}
   \end{adjustbox}
   \label{tab:all}
\end{table*}

\subsection{Main Results}

Table \ref{tab:all} summarizes our main results. As expected, ensembling-based methods (ENS and DiWA) achieve very high accuracy. However, they require much more training computation (see Cost(T)) as they need to train multiple models (18 in the experiment) independently. Our method achieves comparable or even superior performance by training a single model. Additionally, its high swap rate not only enhances generalization but also improves memory efficiency and reduces computational costs during gradient calculation. Note that during inference, High-rate Mixout has the same cost as the ERM baselines (see Cost(I)). For the ``Large Dropout" method, we were unable to reproduce the reported results in~\citet{zhang2024fine}. In their study, they use the SGD optimizer instead of Adam and train for twice as many iterations per dataset compared to the DomainBed default configuration, which is what we use.  This could account for the discrepancies. 

\nbf{TerraIncognita.} Our approach achieves significantly higher performance on TerraIncognita across both architectures compared to other datasets. As noted by~\citet{chen2023explore}, TerraIncognita exhibits the highest feature diversity shift among the DomainBed datasets, reflecting substantial variation in input features across domains. Prior work~\cite{zhang2023learning,rame2022diverse} has shown that richer representations with redundant information, such as those learned by ensemble-based methods, are particularly beneficial in such high-diversity settings. Our method leverages this insight by dynamically swapping pre-trained and fine-tuned parameters at a high rate, fostering the discovery of more robust and diverse feature representations. 

\begin{figure}[t]
    \centering
    \includegraphics[width=1.0\linewidth]{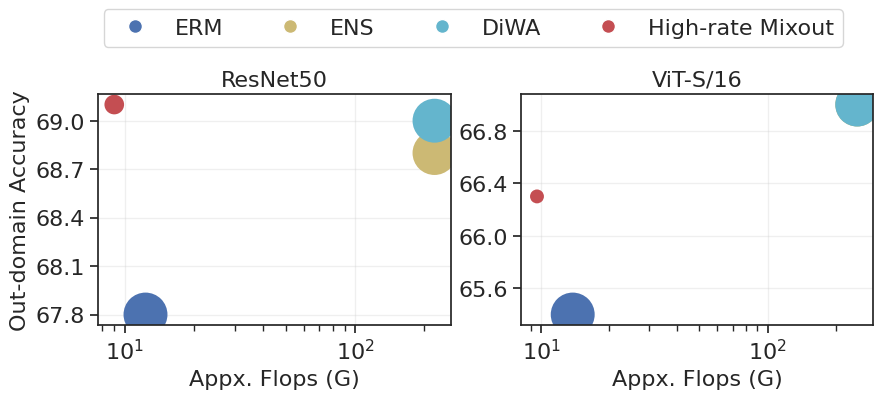}
    \caption{\textbf{Performance versus computational and memory cost for the backward pass across different architectures and methods.} The x-axis is shown on a logarithmic scale. Each bubble area is proportional to the gradient memory usage during the training of a model. High-rate Mixout achieves competitive performance compared to ensemble-based methods while requiring significantly less computation and memory during training.}
    \label{fig:flops}
\end{figure}

\nbf{Computational analysis.} Figure \ref{fig:flops} presents the out-domain performance versus the approximated FLOPs per gradient calculation across five DomainBed benchmarks. For ENS and DiWA, we estimate the FLOPs as the total computation required for training 18 models. We estimate Mixout FLOPs based on the proportion of swapped parameters. While Mixout incurs the same computational overhead as ERM during the forward pass, it achieves efficiency gains in the backward pass by computing gradients for only a subset of parameters. In particular, as shown in Section \ref{sec:ablation}, achieving robustness through Mixout typically requires a swap rate of 0.9 for ViT and 0.8 for ResNet. In other words, this means freezing 90\% of the weights in MLP (CONV) layers, thereby eliminating gradient computations and storage for the pre-trained weights during backpropagation. As a result, the backward pass reduces FLOPs by 45\% (skipping pre-trained weight gradients) and memory by 90\% (no storage for pre-trained gradients). Activation gradient computations remain unchanged, as they depend on all weights (refer to the appendix for the full computation details).

\subsection{Ablation Studies} \label{sec:ablation}

In this section, we provide deeper insights into the underlying mechanism of our proposed method through an ablation study on the OfficeHome dataset. Unless otherwise specified, the experimental setup remains consistent with the main experiments. \\

\begin{figure*}[t]
    \centering
    \includegraphics[width=1.0\linewidth]{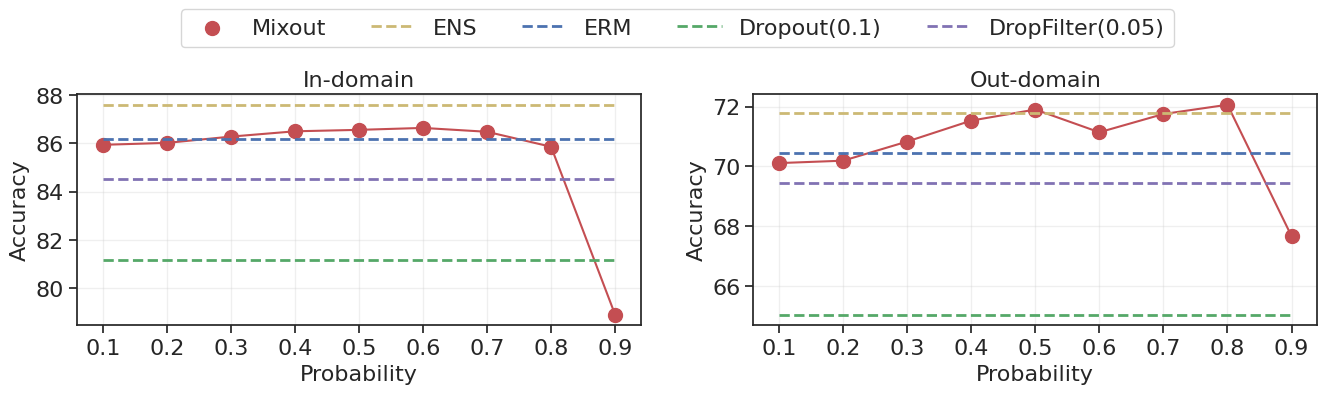}
    \caption{\textbf{Comparison between in-domain and out-domain accuracy of Mixout and Dropout/DropFilter on OfficeHome dataset with ResNet50 architecture.} Unlike ViT-S/16, Mixout with structured masking is better for both the in and out domain performance. We show Dropout and DropFilter results for a probability of 0.1 and 0.05, respectively, as performance rapidly declines to zero for higher probabilities in both cases. Mixout at the probability of 0.0 is equivalent to ERM.}
    \label{fig:r50_iid_ood}
\end{figure*}

\nbf{Effect of structural swapping.} 
For both CNN and ViT architectures, Mixout can be applied with or without structural constraints. Specifically, in ViTs, we can either swap entire outgoing weights from a neuron or randomly replace individual connections. Similarly, in CNNs, instead of replacing parameters individually, we swap entire convolutional kernels within ResNet layers, preserving structural coherence. While both Mixout variants perform similarly for ViTs, we observe a significant performance gap in ResNet50, where the structured version of Mixout proves more effective compared to its unstructured variants (gap of $1.6\%$). This highlights the effectiveness of structured Mixout for CNN in disrupting spatial feature correlations, thereby reducing overfitting and improving generalization. \\

\nbf{Effect of different levels of Mixout.} Figure \ref{fig:r50_iid_ood} extends the analysis in Figure \ref{fig:vit_iid_ood} to ResNet50, evaluating Mixout (with structured masking in this case) performance across swap probabilities (0.1–0.9). For comparison, we also include DropFilter, which disables entire convolutional filters rather than individual neurons. A notable divergence emerges between architectures: While ViT achieves comparable optimal in-domain performance with Mixout and Dropout, ResNet50 exhibits starkly different behavior. Here, Dropout underperforms ERM by a significant margin (in-domain gap of 5.6\% and ood gap of 5.4\% at 0.1 probability), likely due to its susceptibility to unstructured noise in convolutional layers. Although DropFilter reduces this gap, it still does not fully close it. Across both architectures Mixout benefits from increased reliance on pre-trained weights. As the swap rate increases, the model retains more pre-trained parameters, enhancing feature robustness and naturally forming diverse subnetworks, amplifying the ensemble effect without explicit diversity constraints. An interesting observation is that while ViT maintains consistent performance gains at higher swap rates, ResNet50 reaches its optimal performance at a swap rate of 0.8 and experiences a sharp drop at a swap rate of 0.9. This divergence stems from fundamental architectural differences: convolutional layers have significantly fewer filters compared to the number of neurons in MLP layers. As a result, excessive swapping in ResNet50 risks deactivating substantial portions of the neural network during training, leading to undertraining and a notable degradation in performance. \\
\begin{table}[ht]
    \centering
    \caption{\textbf{Out-of-domain accuracy of different techniques with our High-rate Mixout on the Office Home dataset.}}
    \begin{adjustbox}{width=0.47\textwidth}
    \begin{tabular}{ccc|ccc}
        \toprule
        {MA~\cite{arpit2022ensemble}} & {L$_2$-SP~\cite{xuhong2018explicit}} & {LP-FT~\cite{kumar2022fine}} & ERM & High-rate Mixout\\
        
        \midrule
        \multicolumn{5}{c}{\textbf{ResNet50}} \\
        \midrule

        &
        &
        &
        70.46 &
        72.14 \\

        \checkmark &
        &
        &
        71.61 &
        \textbf{72.66} \\

        &
        \checkmark &
        &
        70.62 &
        72.16 \\

        &
        &
        \checkmark &
        70.68 &
        \underline{72.21} \\

        \midrule
        \multicolumn{5}{c}{\textbf{ViT-S/16}} \\
        \midrule

        &
        &
        &
        71.14 &
        \textbf{72.58} \\

        \checkmark &
        &
        &
        71.90 &
        \underline{72.39} \\

        &
        \checkmark &
        &
        71.19 &
        72.49 \\

        &
        &
        \checkmark &
        71.08 &
        72.57 \\

        \bottomrule
    \end{tabular}
    \end{adjustbox}
    \label{tab:abl_ft}
\end{table}

\begin{table}[!htb]
  \centering
\caption{\textbf{Out-of-domain accuracy for PEFT techniques and High-rate Mixout on the Office Home dataset.}}
\begin{adjustbox}{width=0.38\textwidth}
\begin{tabular}{l|cccc}
    \toprule
    {Method} & Cost(T) & Cost(I) & Avg. \\
    \midrule
    \multicolumn{4}{c}{\textbf{ResNet50}} \\
    \midrule
    
    Fixed Mixout &
    .73$\times$        &
    1$\times$  &
    67.25    \\

    High-rate Mixout &
    .73$\times$      &
    1$\times$  &
    \textbf{72.14}    \\

    \midrule
    \multicolumn{4}{c}{\textbf{ViT-S/16}} \\
    \midrule

    LoRA~\cite{hu2021lora} &
    .48$\times$    &
    1$\times$  &
    71.48    \\

    Fixed Mixout &
    .69$\times$        &
    1$\times$  &
    68.92    \\

    High-rate Mixout &
    .69$\times$      &
    1$\times$  &
    \textbf{72.58}    \\

    \bottomrule
    
\end{tabular}
\end{adjustbox}
\label{tab:abl_peft}
\end{table}%

\nbf{Further leveraging pre-trained knowledge.} Table \ref{tab:abl_ft} compares the impact of Mixout when integrated with techniques designed to leverage pre-trained representations against the performance of the same techniques applied to the ERM baseline. This comparison highlights Mixout’s ability to enhance the effectiveness of those techniques by better utilizing pre-trained knowledge. Notable techniques include tuning the classifier head before fine-tuning the entire network (LP-FT)~\cite{kumar2022fine}; Model Averaging (MA), which averages weights during training to improve robustness~\cite{arjovsky2019invariant}; and weight decay towards pre-trained model parameters (L$_2$-SP)~\cite{xuhong2018explicit}. The key observation is that while these methods improve the ERM baselines across both architectures, they still fall short of Mixout's performance. Additionally, while these methods enhanced Mixout performance for ResNet50, they had no significant impact on ViT-S/16. \\

\nbf{Comparison with PEFTs.}
Table~\ref{tab:abl_peft} compares Mixout with two variants of PEFT methods. Fixed Mixout is a simplified version of Mixout that employs a fixed mask over the pre-trained weights throughout training. LoRA~\cite{hu2021lora} reduces the number of trainable parameters by learning low-rank update matrices while keeping the base weights frozen; we use a rank of 64. We exclude LoRA from ResNet50 experiments, as this is not standard practice in the literature. For Fixed Mixout, we start by employing the same swap rate as High-rate Mixout, but observe that the training quickly becomes unstable and starts to diverge. Even after reducing the rate to 0.1, its performance remained inferior to High-rate Mixout, highlighting the necessity of dynamic swapping. While LoRA improves the out-domain performance over ERM, it still underperforms compared to both High-rate Mixout and ENS, emphasizing the value of ensembling effects in our approach. In terms of computation, High-rate Mixout incurs a slightly higher training cost than LoRA but shares the same inference cost as LoRA and ERM: LoRA merges the low-rank updates with the frozen weights, while High-rate Mixout uses rescaled weights at inference. \\

\begin{figure}[t]
    \centering
    \begin{subfigure}[t]{0.46\textwidth}
     \centering
     \includegraphics[width=1.0\textwidth]{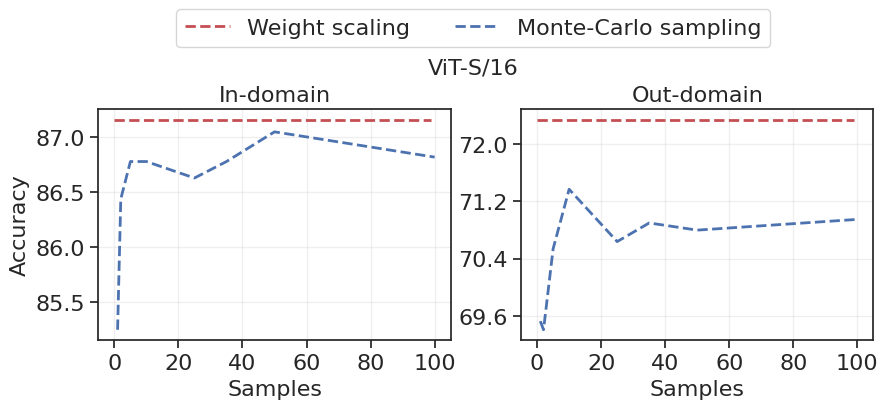}
    \end{subfigure}
    \hfill
    \begin{subfigure}[t]{0.46\textwidth}
        \centering
        \includegraphics[width=1.0\textwidth]{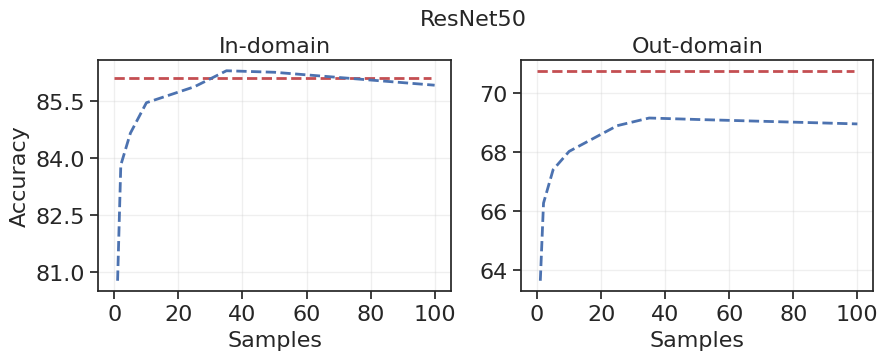}
    \end{subfigure}
    \caption{\textbf{Test accuracy as a function of the number of MC samples $k$ averaged at inference time (blue dashed curves) for a model trained with High-rate Mixout and evaluated on the Art domain from OfficeHome.} For reference, the computationally cheaper single-pass weight scaling approximation is indicated by red horizontal lines.}
    \label{fig:mcmixout}
\end{figure}

\nbf{Monte-Carlo model averaging vs. weight scaling.}
A principled way to obtain the prediction of a model trained with Mixout is to draw $K$ stochastic sub-networks at test time and average their outputs, i.e.\ Monte-Carlo (MC) model averaging. In the limit $K \to \infty$, this recovers the exact ensemble, but the computational cost grows linearly with $K$. \cref{fig:mcmixout} explores this accuracy–efficiency trade-off for the High-rate Mixout model evaluated on the unseen Art domain. For both backbones, the MC curve rises steeply: 5–10 samples close most of the gap to the inexpensive \emph{weight-scaling} surrogate, and roughly 50 samples are sufficient to match it on the training domains, after which returns saturate. The picture changes under a domain shift. On the unseen domain from the OfficeHome dataset, the weight-scaled network remains about one percentage point more accurate even after averaging 100 sub-nets. We hypothesise that unseen-domain images activate feature pathways that are \emph{sparsely} selected during training; every MC draw therefore zeroes out some critical weights—but rarely co-activated—on the source domains. Averaging many such sub-nets consequently produces logits biased toward features that survive more often, leaving these rare yet important pathways under-represented. Weight scaling, in contrast, retains \emph{all} parameters and thus evaluates a deterministic ``mean network'' whose smoother activations exhibit lower variance and preserve those infrequent but crucial features. The result is a small but persistent robustness edge for weight scaling when generalising beyond the training distribution. \\

\nbf{Functional similarity of sub-networks.} 
Ensemble gains hinge not only on the strength of the individual models but also on their functional diversity~\cite{dietterich2000ensemble}.
We quantify this diversity for High-rate Mixout by the \emph{pairwise disagreement rate} $\frac{1}{N}\sum_{n=1}^N \Bigl( f(x_n; \theta_i) - f(x_n; \theta_j) \Bigr)$, where $f(x;\theta)$ is the predicted class label of subnet $i$ or $j$ for input $x$. Using 50 randomly drawn subnet pairs of a ViT-S/16 backbone, we obtain an average disagreement of 0.28 on the source domains and 0.43 on the unseen \textit{Art} domain from OfficeHome. The higher out-domain value indicates that High-rate Mixout generates more diverse decision functions under distribution shift, enabling the model to exploit complementary errors and achieve more robustness.

\section{Conclusion}

In DG benchmarks, ensemble-based methods consistently outperform conventional DG approaches by producing richer and more robust representations. These methods leverage pre-trained models and the stochastic nature of training to enhance generalization across diverse domains. However, their superior performance comes at a substantial computational cost, as fully realizing ensemble benefits typically requires training and storing multiple models with different initializations and hyperparameters.
In this study, we introduce High-rate Mixout, a regularization technique that achieves ensemble-level performance without the heavy computational burden of full ensembles. Our approach dynamically swaps a large portion of fine-tuned parameters (90\% for ViTs, 80\% for ResNets) with their pre-trained counterparts, effectively constraining the model near its pre-trained weights while still promoting adaptation. This high-rate, dynamic swapping uncovers diverse subnetworks throughout training—a key driver of ensemble-like robustness—while significantly reducing backward computation (by up to 45\%) and gradient memory usage (by up to 90\%).
Importantly, we adapt Mixout for convolutional architectures by swapping entire convolutional kernels, addressing the spatial correlation challenges seen with unstructured masking. Extensive ablation studies on ResNet50 and ViT-S/16 backbones validate these design choices. Our empirical results on the DomainBed benchmark show that High-rate Mixout achieves out-of-domain accuracy comparable to ensemble-based methods while substantially reducing training costs.

{\small
\bibliographystyle{ieeenat_fullname}
\bibliography{main}
}

\newpage
\onecolumn
\appendix

\section{Computational and Memory Analysis}

We analyze memory usage and FLOPs for each method, where total FLOPs per iteration include forward and backward passes. The backward pass computes weight gradients ($\nabla W$) and input gradients ($\nabla X$), each requiring $\sim\!1\times$ the FLOPs of the forward pass. 

\subsection{Single-Run Methods}
For conventional methods (e.g., ERM):
\[
\text{FLOPs} = 3 \times \text{Forward FLOPs} =
\begin{cases}
12.3\,\text{G} & (\text{ResNet50: } 4.1 \times 3), \\
13.8\,\text{G} & (\text{ViT-S/16: } 4.6 \times 3).
\end{cases}
\]
We need to store 100\% of weight and activation gradients.

\subsection{Ensemble-Based Methods}
For 18-model ensembles:
\[
\text{FLOPs} = 18 \times \text{ERM FLOPs}, \quad \text{Memory} = \text{ERM Memory} \, (\text{sequential training}).
\]

\subsection{Mixout}
For swap rate $p\%$:
\begin{itemize}
    \item \textbf{FLOPs}: Skips $p\%$ of $\nabla W$ computation:
    \[
    \text{FLOPs} = \text{Base} \times \left(2 + \frac{p}{100}\right), \quad \text{where } \text{Base} = \text{Forward FLOPs}.
    \]
    \item \textbf{Memory}: Reduces storage by $p\%$:
    \[
    \text{Memory} = \left(1 - \frac{p}{100}\right) \times \text{Full Memory}.
    \]
    Input gradient computations remain unchanged.
\end{itemize}

\subsection{LoRA for ViT-S/16}

\noindent\textbf{Forward Pass:}
Replacing all linear layers in ViT-S/16 with LoRA of rank 64 adds the following FLOPs:
\[
\text{FLOPs}_{\text{LoRA-forward}} = 12 \cdot N \cdot \left[ 4 \cdot (D \cdot r + r \cdot D) + 2 \cdot (D \cdot r + r \cdot 4D) \right],
\]
where \(N = 196\) (sequence length), \(D = 384\) (hidden dim), \(4D = 1536\) (MLP intermediate dim), and \(r = 64\). Substituting values:
\[
\text{FLOPs}_{\text{LoRA-forward}} = 12 \cdot 196 \cdot \left[ 4 \cdot (384 \cdot 64 + 64 \cdot 384) + 2 \cdot (384 \cdot 64 + 64 \cdot 1536) \right] = 1.04\,\text{GFLOPs}.
\]
\textbf{Total forward FLOPs}:
\[
4.6\,\text{GFLOPs} + 1.04\,\text{GFLOPs} = \mathbf{5.64\,\text{GFLOPs}}.
\]

\noindent\textbf{Backward Pass:}
LoRA backward FLOPs are twice the forward FLOPs (gradient computations for \(A\) and \(B\)):
\[
\text{FLOPs}_{\text{LoRA-backward}} = 2 \cdot \text{FLOPs}_{\text{LoRA-forward}} = 2.08\,\text{GFLOPs}.
\]
Final FLOPs is 
\[
9.2\,\text{GFLOPs} - 4.6\,\text{GFLOPs} + 2.08\,\text{GFLOPs} = \mathbf{6.68\,\text{GFLOPs}}.
\]

\section{Implementation details} \label{apx:impl}

Following DomainBed benchmark \cite{gulrajani2020search}, we evaluate our method on five diverse datasets. PACS~\cite{li2017deeper}, VLCS \cite{fang2013unbiased}, Office Home~\cite{venkateswara2017deep}, TerraIncognita~\cite{beery2018recognition} and DomainNet~\cite{peng2019moment}. We report out-of-domain accuracies for each domain and their average using a leave-one-out cross-validation method. We use Adam~\cite{kingma2014adam} optimizer with a mini-batch containing all domains and $32$ examples per domain. We follow~\cite{cha2021swad} and train models for $15000$ steps on DomainNet and $5000$ steps for other datasets, corresponding to a variable number of epochs dependent on dataset size. Every experiment is repeated three times with different seeds. We leave 20\% of the source domain data for validation. For each domain within a dataset, we follow the procedure described in ~\cite{cha2021swad}, sampling $18$ hyperparameter sets from the search space and fine-tuning ERM to identify the best configuration. We then conduct a separate hyperparameter search tailored to each method. Each domain is sequentially used as the target (test) domain, while the remaining domains are utilized as source (training) domains (please refer to the appendix for more details on datasets and hyperparameter configurations). 

\begin{table}[h]
    \centering
  \caption{Hyperparameters used for all methods in and their respective distributions for grid search.}
    \begin{tabular}{l@{\hskip 0.1in}c}
        \toprule
        \textbf{Hyperparameter} & \textbf{Search Space} \\
        \addlinespace[1pt]
        \midrule
        batch size &
        32 \\
        learning rate &
        \{1e-5, 3e-5, 5e-5\} \\
        classifier dropout &
        \{0.0, 0.1, 0.5\} \\
        weight decay &
        \{1e-4, 1e-6\} \\
        \midrule
        swap rate &
        \{0.1, 0.2, $\cdots$, 0.9 \} \\

        \bottomrule
    \end{tabular}
  \label{tab:hyp}
\end{table}

\subsection{Datasets} \label{apex:dataset}

\paragraph{\textbf{PACS:}}is a 7-way object classification task with $4$ domains: art, cartoon, photo, and sketch, with $9,991$ samples~\cite{li2017deeper}. 

\paragraph{\textbf{VLCS:}} is a 5-way classification task from $4$ domains: Caltech101, LabelMe, SUN09, and VOC2007. There are $10,729$ samples. This dataset mostly contains real photos. The distribution shifts are subtle and simulate real-life scenarios well~\cite{fang2013unbiased}. 

\paragraph{\textbf{Office Home:}} is a 65-way classification task depicting everyday objects from $4$ domains: art, clipart, product, and real, with a total of $15,588$ samples~\cite{venkateswara2017deep}. 

\paragraph{\textbf{TerraIncognita:}} is a 10-way classification problem of animals in wildlife cameras, where the $4$ domains are different locations, L100, L38, L43, L46. There are $24,788$ samples. This represents a realistic use case where generalization is indeed critical~\cite{beery2018recognition}. 

\paragraph{\textbf{DomainNet:}} is a 345-way object classification task from 6 domains: clipart, infograph, painting, quickdraw, real, and sketch. With a total of $586,575$ samples, it is larger than most of the other evaluated datasets in both samples and classes~\cite{peng2019moment}. 

\section{Full Results} \label{apx:full_res}

In this section, we show detailed results of Table 1 of the main manuscript. Tables \ref{tab:pacs}, \ref{tab:vlcs}, \ref{tab:officehome}, \ref{tab:terra} \ref{tab:domain_net} show full results on PACS, VLCS, OfficeHome, TerraIncognita, and DomainNet datasets, respectively. The provided tables summarize the obtained out-of-distribution accuracy for every domain within the five datasets. Standard deviations are reported with different seeds when possible. To guarantee the comparability of the results, we followed the same experimental setting as in DomainBed \cite{gulrajani2020search}.

\begin{table*}[p]
    \centering
    \caption{OOD accuracies on PACS.}
    \begin{adjustbox}{width=0.7\textwidth}
    \begin{tabular}{l|ccccc}
        \toprule
        {Method} & A & C & P & S & \textbf{Avg.}\\
        
        \midrule
        \multicolumn{6}{c}{\textbf{ResNet50}} \\
        \midrule
        \multicolumn{6}{l}{\textbf{Multi-run training (18 Models)}} \\
        \midrule

        ENS &
        90.85  &
        83.53  &
        98.88  &
        82.95  &
        89.05 \\

        DiWA &
        92.01  &
        84.01  &
        99.18  &
        81.65  &
        89.21 \\

        \midrule
        \multicolumn{6}{l}{\textbf{Single-run training}} \\
        \midrule

        CORAL &
        89.36$\scriptstyle\pm 0.76$  &
        80.44$\scriptstyle\pm 0.99$  &
        98.58$\scriptstyle\pm 0.11$  &
        81.23$\scriptstyle\pm 0.82$  &
        87.40$\scriptstyle\pm 0.67$    \\  

        Large Dropout &
        87.78$\scriptstyle\pm 1.31$  &
        82.68$\scriptstyle\pm 0.28$  &
        98.43$\scriptstyle\pm 0.15$  &
        79.66$\scriptstyle\pm 0.75$  &
        87.14$\scriptstyle\pm 0.62$    \\  

        ERM &
        90.81$\scriptstyle\pm 0.87$  &
        81.68$\scriptstyle\pm 0.78$  &
        98.68$\scriptstyle\pm 0.26$  &
        79.45$\scriptstyle\pm 0.94$  &
        87.66$\scriptstyle\pm 0.71$    \\    

        \textbf{High-rate Mixout} &
        91.07$\scriptstyle\pm 0.14$  &
        83.46$\scriptstyle\pm 0.18$  &
        99.15$\scriptstyle\pm 0.04$  &
        79.32$\scriptstyle\pm 0.50$  &
        88.25$\scriptstyle\pm 0.22$    \\

        \midrule
        \multicolumn{6}{c}{\textbf{ViT-S/16}} \\
        \midrule
        \multicolumn{6}{l}{\textbf{Multi-run training (18 Models)}} \\
        \midrule

        ENS &
        90.30  &
        82.20  &
        99.10  &
        79.96  &
        87.89    \\

        DiWA &
        90.79  &
        83.00  &
        99.25  &
        80.47  &
        88.38   \\

        \midrule
        \multicolumn{6}{l}{\textbf{Single-run training}} \\
        \midrule

        ERM &
        87.27$\scriptstyle\pm 0.60$  &
        81.11$\scriptstyle\pm 0.96$  &
        98.30$\scriptstyle\pm 0.11$  &
        77.47$\scriptstyle\pm 1.03$  &
        86.04$\scriptstyle\pm 0.68$    \\
        
        \textbf{High-rate Mixout} &
        89.91$\scriptstyle\pm 0.47$  &
        80.99$\scriptstyle\pm 0.37$  &
        98.83$\scriptstyle\pm 0.05$  &
        77.42$\scriptstyle\pm 1.21$  &
        86.79$\scriptstyle\pm 0.52$    \\

        \bottomrule
    \end{tabular}
    \end{adjustbox}
   \label{tab:pacs}
\end{table*}

\begin{table*}[p]
    \centering
    \caption{OOD accuracies on VLCS.}
    \begin{adjustbox}{width=0.7\textwidth}
    \begin{tabular}{l|ccccc}
        \toprule
        {Method} & C & L & S & V & \textbf{Avg.}\\
        
        \midrule
        \multicolumn{6}{c}{\textbf{ResNet50}} \\
        \midrule
        \multicolumn{6}{l}{\textbf{Multi-run training (18 Models)}} \\
        \midrule

        ENS &
        98.06  &
        64.89  &
        76.28  &
        80.90  &
        80.03 \\

        DiWA &
        98.06  &
        63.67  &
        76.96  &
        89.64  &
        79.83 \\

        \midrule
        \multicolumn{6}{l}{\textbf{Single-run training}} \\
        \midrule

        CORAL &
        98.82$\scriptstyle\pm 0.10$  &
        64.94$\scriptstyle\pm 0.69$  &
        76.83$\scriptstyle\pm 0.77$  &
        79.46$\scriptstyle\pm 0.52$  &
        80.01$\scriptstyle\pm 0.52$    \\  

        Large Dropout &
        97.76$\scriptstyle\pm 0.46$  &
        64.82$\scriptstyle\pm 0.35$  &
        74.41$\scriptstyle\pm 0.38$  &
        80.25$\scriptstyle\pm 0.54$  &
        79.31$\scriptstyle\pm 0.43$    \\

        ERM &
        98.06$\scriptstyle\pm 0.15$  &
        64.28$\scriptstyle\pm 0.49$  &
        76.72$\scriptstyle\pm 0.48$  &
        79.48$\scriptstyle\pm 0.60$  &
        79.64$\scriptstyle\pm 0.43$    \\  

        \textbf{High-rate Mixout} &
        98.20$\scriptstyle\pm 0.10$  &
        65.68$\scriptstyle\pm 0.12$  &
        73.88$\scriptstyle\pm 0.55$  &
        79.85$\scriptstyle\pm 0.80$  &
        79.40$\scriptstyle\pm 0.39$    \\

        \midrule
        \multicolumn{6}{c}{\textbf{ViT-S/16}} \\
        \midrule
        \multicolumn{6}{l}{\textbf{Multi-run training (18 Models)}} \\
        \midrule

        ENS &
        97.26  &
        65.65  &
        77.53  &
        83.38  &
        80.96    \\

        DiWA &
        96.64  &
        64.85  &
        77.68  &
        82.90  &
        80.52    \\

        \midrule
        \multicolumn{6}{l}{\textbf{Single-run training}} \\
        \midrule

        ERM &
        96.91$\scriptstyle\pm 0.14$  &
        64.49$\scriptstyle\pm 0.22$  &
        75.65$\scriptstyle\pm 0.93$  &
        82.29$\scriptstyle\pm 0.13$  &
        79.83$\scriptstyle\pm 0.36$    \\

        \textbf{High-rate Mixout} &
        95.88$\scriptstyle\pm 0.46$  &
        64.74$\scriptstyle\pm 0.46$  &
        76.96$\scriptstyle\pm 0.20$  &
        79.49$\scriptstyle\pm 0.57$  &
        79.27$\scriptstyle\pm 0.42$    \\

        \bottomrule
    \end{tabular}
    \end{adjustbox}
   \label{tab:vlcs}
\end{table*}

\begin{table*}[p]
    \centering
    \caption{OOD accuracies on OfficeHome.}
    \begin{adjustbox}{width=0.7\textwidth}
    \begin{tabular}{l|ccccc}
        \toprule
        {Method} & A & C & P & R & \textbf{Avg.}\\
        
        \midrule
        \multicolumn{6}{c}{\textbf{ResNet50}} \\
        \midrule
        \multicolumn{6}{l}{\textbf{Multi-run training (18 Models)}} \\
        \midrule

        DiWA &
        70.55  &
        53.64  &
        79.76  &
        83.02  &
        71.74 \\

        ENS &
        69.77  &
        54.04  &
        79.95  &
        83.33  &
        71.77 \\

        \midrule
        \multicolumn{6}{l}{\textbf{Single-run training}} \\
        \midrule

        CORAL &
        70.08$\scriptstyle\pm 0.50$  &
        53.20$\scriptstyle\pm 0.39$  &
        78.95$\scriptstyle\pm 0.29$  &
        82.69$\scriptstyle\pm 0.18$  &
        71.23$\scriptstyle\pm 0.34$    \\

        Large Dropout &
        68.64$\scriptstyle\pm 0.74$  &
        53.30$\scriptstyle\pm 0.30$  &
        78.13$\scriptstyle\pm 0.43$  &
        82.56$\scriptstyle\pm 0.05$  &
        70.66$\scriptstyle\pm 0.38$    \\   

        ERM &
        68.95$\scriptstyle\pm 1.16$  &
        52.13$\scriptstyle\pm 0.67$  &
        78.61$\scriptstyle\pm 0.48$  &
        82.14$\scriptstyle\pm 0.49$  &
        70.46$\scriptstyle\pm 0.70$    \\

        \textbf{High-rate Mixout} &
        71.06$\scriptstyle\pm 0.68$  &
        54.20$\scriptstyle\pm 0.35$  &
        79.99$\scriptstyle\pm 0.11$  &
        83.30$\scriptstyle\pm 0.06$  &
        72.14$\scriptstyle\pm 0.30$    \\

        \midrule
        \multicolumn{6}{c}{\textbf{ViT-S/16}} \\
        \midrule
        \multicolumn{6}{l}{\textbf{Multi-run training (18 Models)}} \\
        \midrule

        ENS &
        71.83  &
        56.10  &
        81.42  &
        82.62  &
        72.99   \\

        DiWA &
        72.40  &
        55.87  &
        81.17  &
        82.36  &
        72.95    \\

        \midrule
        \multicolumn{6}{l}{\textbf{Single-run training}} \\
        \midrule

        ERM &
        68.74$\scriptstyle\pm 0.44$  &
        54.16$\scriptstyle\pm 0.61$  &
        80.01$\scriptstyle\pm 0.27$  &
        81.64$\scriptstyle\pm 0.20$  &
        71.14$\scriptstyle\pm 0.38$    \\

        \textbf{High-rate Mixout} &
        72.11$\scriptstyle\pm 0.10$  &
        55.05$\scriptstyle\pm 0.32$  &
        80.78$\scriptstyle\pm 0.11$  &
        82.36$\scriptstyle\pm 0.18$  &
        72.58$\scriptstyle\pm 0.18$    \\

        \bottomrule
    \end{tabular}
    \end{adjustbox}
   \label{tab:officehome}
\end{table*}

\begin{table*}[p]
    \centering
    \caption{OOD accuracies on TerraIncognita.}
    \begin{adjustbox}{width=0.7\textwidth}
    \begin{tabular}{l|ccccc}
        \toprule
        {Method} & L100 & L38 & L43 & L46 & \textbf{Avg.}\\
        
        \midrule
        \multicolumn{6}{c}{\textbf{ResNet50}} \\
        \midrule
        \multicolumn{6}{l}{\textbf{Multi-run training (18 Models)}} \\
        \midrule

        ENS &
        63.67  &
        46.44  &
        63.48  &
        42.83  &
        54.10 \\

        DiWA &
        62.98  &
        50.44  &
        62.47  &
        46.85  &
        55.68 \\

        \midrule
        \multicolumn{6}{l}{\textbf{Single-run training}} \\
        \midrule

        CORAL &
        58.21$\scriptstyle\pm 1.94$  &
        47.59$\scriptstyle\pm 1.62$  &
        57.65$\scriptstyle\pm 0.51$  &
        38.98$\scriptstyle\pm 1.84$  &
        50.61$\scriptstyle\pm 1.48$    \\

        ERM &
        59.53$\scriptstyle\pm 2.79$  &
        48.93$\scriptstyle\pm 1.79$  &
        61.87$\scriptstyle\pm 1.57$  &
        40.13$\scriptstyle\pm 3.17$  &
        52.62$\scriptstyle\pm 2.33$    \\  

        Large Dropout &
        61.31$\scriptstyle\pm 2.79$  &
        47.65$\scriptstyle\pm 1.99$  &
        60.71$\scriptstyle\pm 0.64$  &
        39.41$\scriptstyle\pm 0.77$  &
        52.27$\scriptstyle\pm 1.55$    \\ 

        \textbf{High-rate Mixout} &
        65.53$\scriptstyle\pm 0.49$  &
        56.93$\scriptstyle\pm 1.32$  &
        64.27$\scriptstyle\pm 0.21$  &
        46.95$\scriptstyle\pm 0.64$  &
        58.42$\scriptstyle\pm 0.66$    \\
        
        \midrule
        \multicolumn{6}{c}{\textbf{ViT-S/16}} \\
        \midrule
        \multicolumn{6}{l}{\textbf{Multi-run training (18 Models)}} \\
        \midrule

        ENS &
        50.49  &
        30.25  &
        54.44  &
        40.24  &
        43.86 \\

        DiWA &
        52.10  &
        30.70  &
        56.52  &
        41.05  &
        45.09 \\

        \midrule
        \multicolumn{6}{l}{\textbf{Single-run training}} \\
        \midrule

        ERM &
        53.66$\scriptstyle\pm 2.48$  &
        27.49$\scriptstyle\pm 2.25$  &
        51.30$\scriptstyle\pm 0.05$  &
        36.75$\scriptstyle\pm 0.04$  &
        42.30$\scriptstyle\pm 1.20$    \\

        \textbf{High-rate Mixout} &
        58.62$\scriptstyle\pm 1.03$  &
        31.71$\scriptstyle\pm 0.35$  &
        56.19$\scriptstyle\pm 0.60$  &
        42.14$\scriptstyle\pm 0.46$  &
        47.16$\scriptstyle\pm 0.61$    \\

        \bottomrule
    \end{tabular}
    \end{adjustbox}
   \label{tab:terra}
\end{table*}

\begin{table*}[p]
    \centering
    \caption{OOD accuracies on DomainNet.}
    \begin{adjustbox}{width=0.9\textwidth}
    \begin{tabular}{l|ccccccc}
        \toprule
        {Method} & C & I & P & Q & R & S & \textbf{Avg.}\\
        
        \midrule
        \multicolumn{8}{c}{\textbf{ResNet50}} \\
        \midrule
        \multicolumn{8}{l}{\textbf{Multi-run training (18 Models)}} \\
        \midrule

        ENS &
        68.66  &
        25.39  &
        56.99  &
        14.58  &
        71.28  &
        57.74  &
        49.11    \\  

        DiWA &
        66.69  &
        25.15  &
        56.73  &
        14.66  &
        70.40  &
        56.79  &
        48.40    \\  

        \midrule
        \multicolumn{8}{l}{\textbf{Single-run training}} \\
        \midrule

        CORAL &
        66.89$\scriptstyle\pm 0.20$  &
        24.43$\scriptstyle\pm 0.20$  &
        54.50$\scriptstyle\pm 0.26$  &
        13.82$\scriptstyle\pm 0.27$  &
        68.34$\scriptstyle\pm 0.31$  &
        56.02$\scriptstyle\pm 0.48$  &
        47.33$\scriptstyle\pm 0.29$    \\  

        Large Dropout &
        67.04$\scriptstyle\pm 0.10$  &
        25.14$\scriptstyle\pm 0.28$  &
        54.48$\scriptstyle\pm 0.11$  &
        13.37$\scriptstyle\pm 0.26$  &
        68.22$\scriptstyle\pm 0.15$  &
        55.90$\scriptstyle\pm 0.12$  &
        47.36$\scriptstyle\pm 0.17$    \\

        ERM &
        67.09$\scriptstyle\pm 0.10$  &
        25.58$\scriptstyle\pm 0.32$  &
        56.21$\scriptstyle\pm 0.97$  &
        14.85$\scriptstyle\pm 0.32$  &
        69.56$\scriptstyle\pm 0.52$  &
        57.61$\scriptstyle\pm 0.62$  &
        48.48$\scriptstyle\pm 0.48$    \\  

        \textbf{High-rate Mixout} &
        66.78$\scriptstyle\pm 0.30$  &
        24.26$\scriptstyle\pm 0.21$  &
        54.90$\scriptstyle\pm 0.26$  &
        14.24$\scriptstyle\pm 0.28$  &
        69.13$\scriptstyle\pm 0.18$  &
        56.84$\scriptstyle\pm 0.36$  &
        47.69$\scriptstyle\pm 0.26$    \\

        \midrule
        \multicolumn{8}{c}{\textbf{ViT-S/16}} \\
        \midrule
        \multicolumn{8}{l}{\textbf{Multi-run training (18 Models)}} \\
        \midrule

        ENS &
        68.76  &
        25.82  &
        57.51  &
        16.47  &
        70.18  &
        56.70  &
        49.24    \\

        DiWA &
        66.76  &
        25.87  &
        56.56  &
        16.54  &
        68.62  &
        55.95  &
        48.38    \\

        \midrule
        \multicolumn{8}{l}{\textbf{Single-run training}} \\
        \midrule

        ERM &
        66.62$\scriptstyle\pm 0.10$  &
        24.85$\scriptstyle\pm 0.14$  &
        55.04$\scriptstyle\pm 0.12$  &
        15.29$\scriptstyle\pm 0.35$  &
        68.53$\scriptstyle\pm 0.20$  &
        54.60$\scriptstyle\pm 0.16$  &
        47.49$\scriptstyle\pm 0.18$    \\

        \textbf{High-rate Mixout} &
        67.48$\scriptstyle\pm 0.11$  &
        23.68$\scriptstyle\pm 0.15$  &
        54.75$\scriptstyle\pm 0.19$  &
        15.68$\scriptstyle\pm 0.31$  &
        68.16$\scriptstyle\pm 0.17$  &
        54.96$\scriptstyle\pm 0.15$  &
        47.45$\scriptstyle\pm 0.12$    \\

        \bottomrule
    \end{tabular}
    \end{adjustbox}
   \label{tab:domain_net}
\end{table*}

\end{document}